\relax
\documentclass[letterpaper]{article} 
\usepackage{aaai21}  
\usepackage{times}  
\usepackage{helvet} 
\usepackage{courier}  
\usepackage[hyphens]{url}  
\usepackage{graphicx} 
\usepackage{natbib}  
\usepackage{caption} 
\usepackage[switch]{lineno}  %
\usepackage{amsmath,amsfonts,amssymb,amsthm}
\theoremstyle{assumption}
\newtheorem{assumption}{Assumption}[section]
\theoremstyle{definition}
\newtheorem{definition}{Definition}[section]

\usepackage[ruled,vlined,norelsize]{algorithm2e}
\makeatletter
\newcommand{\removelatexerror}{\let\@latex@error\@gobble}
\usepackage{color}

\makeatother
\usepackage{subcaption}
\title{Improving Federated Relational Data Modeling via Basis Alignment \\and Weight Constraint}
\author{

    Yilun Lin, 
    Chaochao Chen,
    Cen Chen,
    Li Wang\\
}
\affiliations{

    Ant Group, No. 556 Xixi Road, Hangzhou, China\\
    \{yilun.lyl, chaochao.ccc, chencen.cc, raymond.wangl\}@antgroup.com

}

\begin{document}
\maketitle

\begin{abstract}
Federated learning (FL) has attracted increasing attention in recent years. As a privacy-preserving collaborative learning paradigm, it enables a broader range of applications, especially for computer vision and natural language processing tasks. 
However, to date, there is limited research of federated learning on relational data, namely Knowledge Graph (KG). 
In this work, we present a modified version of the graph neural network algorithm that performs federated modeling over KGs across different participants. Specifically, to tackle the inherent \textit{data heterogeneity} issue and inefficiency in \textit{algorithm convergence}, we propose a novel optimization algorithm, named FedAlign, with 1) optimal transportation (OT) for on-client personalization and 2) weight constraint to speed up the convergence.  Extensive experiments have been conducted on several widely used datasets. Empirical results show that our proposed method outperforms the state-of-the-art FL methods, such as FedAVG and FedProx, with better convergence.
\end{abstract}

\section{Introduction\label{sec-intro}}

Relational data organizes factual knowledge that is valuable for a wide range of applications~\cite{Ji-2020}, such as question answering~\cite{Seyler-2015,He-2017} and information retrieval~\cite{Dalton-2014,Xiong-2015a}.
As most of the knowledge bases in the real world are incomplete, predicting missing information in knowledge bases, i.e.,  statistical relational learning (SRL) \cite{Getoor-2007} task, has attracted great attention from both academia and industry.

An effective way for solving SRL tasks is to utilize graph neural network (GNN)~\cite{Scarselli-2009}, such as relational graph convolutional network (RGCN)~\cite{Schlichtkrull-2017}.
However, data are usually scattered in different companies and institutions, especially for the financial domain where data are sensitive by nature. Collecting data from these institutions is difficult or even forbidden by regulation. 
Privacy-preserving SRL methods that allow secure collaborative training among different participants are still less studied, thus, hinders the wider applications of graph modeling. 
A promising direction for such collaborative training purpose is to explore Federated Learning (FL).
The first algorithm proposed by Google is the FedAVG. It jointly learns a global model with multiple data sources by only exchanging the gradients or model parameters while keeping the raw data stay locally, thus limits the possibility of information leakage~\cite{McMahan-2017}.

Although FL has achieved significant progress and has been widely applied, the study on combining FL and graph learning based methods remains less explored. 
One of the possible reasons might be the inherent heterogeneity for different graph datasets. 
Such graph heterogeneity exists either in the statistical sense that the number of nodes and edges could be extremely varied for graph datasets even with similar types of nodes and edges, or in a structural way where an entity in separate graph datasets might be identical but with different neighborhoods. 
However, current FL algorithms typically assume IID training data to perform well, the fundamental heterogeneity problem in federated graphs might seriously degrade the performance of the jointly trained model.

The recently proposed FedProx \cite{Li-2020} tackles the systems and statistical heterogeneity in federated networks by adding a penalty factor on local loss function to encourage the local model closer to the global one. However, such a method can not be applied to graph related models directly, 
as the local data in graph modeling tasks has a different property as in batched data, and hence models diverge more significantly. One of the most vital differences is that, the batched data can be separated into batches during the training process, while the graph data can not be split at all. Such difference leads to a high variance of local weight and difficulties in aligning the local and global models.
Hence, a more sophisticated alignment method needs to be designed. 

In this paper, we first review the possible reasons why local models may differ significantly in the federated setting. 
We find that the non-separability of graph data and the complex graph model design may aggravate local model divergence.
Based on these insights, we propose a simple yet effective solution, called FedAlign, by constraining the loss function to be $L$-Lipschitz smooth and measuring the optimal transportation (OT) distance of hidden layers.
Extensive experiments are conducted on several public datasets. Results show that the proposed FedAlign outperforms the state-of-art federated learning methods, such as FedAVG and FedProx on modeling relational data. 

We summarize our main contributions as follows: 
\begin{itemize}
    \item 1) We first propose how to build a naive federated RGCN model based on relational data, by directly using the existing FL technique. 
    \item 2) We then review the problems (objective divergence and unsmoothness) of the naive solution and propose an advanced method using basis alignment and weight constraint. 
    \item 3) We finally conduct experiments on three benchmark datasets and the results demonstrate the effectiveness of our proposed method.
\end{itemize}



\section{Preliminaries}\label{sec-preliminaries}
In this section, we present the preliminaries on federated learning and graph neural network, so as to propose a modified version of graph neural network for modeling the relational data over the federated networks in the later section.

\subsection{Federated Learning}

Federated Learning (FL) is an emerging technique that aims to preserve privacy and boost model performance on edge devices. A typical use case of FL is to train a keyboard prediction model on the mobile phone, which predicts the next word according to the last input of a device user. The inputs on the mobile phone are highly private and the user will be reluctant to share the data to the server.
However, the limited data on a single phone can hardly be enough for model training. 
FL was first proposed by~\citet{McMahan-2017} to solve such a problem by adopting a collaboratively training paradigm without sharing local data.

Federated Averaging (FedAVG) is the most commonly used algorithm in FL~\cite{McMahan-2017}.
In this algorithm, each client $k \in K$ (i.e., the local device such as a mobile phone of a user) updates its local model using data collected on itself.  The sever chooses clients periodically, collects the parameters $w_{t}^{k}$ and  aggregates them to compute the global parameters as follows:
\begin{equation}
	w_{t+1} \leftarrow \sum_{k=1}^{K} \frac{n_{k}}{n} w_{t+1}^{k}, \label{eq-fedavg-agg}
\end{equation}
where $n_k$ is the data size in each client $k$, and $n$ is the size of all the data used in this update round. Finally, clients replace their local parameters with the global ones.

To prevent gradients from leaking sensitive information during the federated optimization, FedAVG can be easily adapted with privacy-preserving techniques, such as differential privacy and secure multiparty communication. 
However, in the real-world federated applications,
the edge devices are not always online and the local data typically varies from device to device. The system and statistical heterogeneity naturally exist.
FedProx~\cite{Li-2020} was proposed recently to especially tackle the system (caused by unreliable communication or stragglers) and statistical heterogeneity (caused by the nature of data collectors).
It has achieved the state-of-art performances over many federated benchmarks. 
The key idea is to introduce a proximal term that limits local updates from diverging:
\begin{equation}
	\min _{w} F_{k}(w)+\frac{\mu}{2} \| w- w^{t} \|^{2}.
	\label{eq-fedprox}
\end{equation}
$F_{k}(w):=\mathbb{E}_{x_{k} \sim \mathcal{D}_{k}}\left[f\left(w ; x_{k}\right)\right]$ in \eqref{eq-fedprox} is the local  counterpart of objective $f$ in the (ideally existing) global task. 
It measures the local empirical risk over possibly differing data distributions $D_k$.
$w^{t}$ denotes the global model and $w$ refers to the local one. 
FedProx can be viewed as a generalization of FedAVG when $\mu=0$.

\subsection{Graph Neural Networks}

Graph neural network (GNN) has shown to be effective in modeling graph data and has achieved state-of-the-art performances on several tasks \cite{Wu-2020,Liu-2018a}. A typical GNN model consists of two parts, i.e., an embedding layer that encodes the graph into learnable vectors and hidden layers that transforms the embedding into task-specific outputs.
Specifically, GNN usually defines a differentiable message-passing function on local neighborhoods\cite{Gilmer-2017}, i.e., : 
\begin{equation}
	h_{i}^{(l+1)}=\sigma\left(\sum_{m \in \mathcal{M}_{i}} g_{m}\left(h_{i}^{(l)}, h_{j}^{(l)}\right)\right) \label{eq-gnn},
\end{equation}
where $h_{i}^{(l+1)} \in \mathcal{R}^{d^{(l)}}$ denotes the hidden state of node $v_i$ in the $l$-layer, with $d^{l}$ being the dimensionality of this layer. 
A message-passing operator $g_m( \cdot , \cdot )$  chooses from the set of incoming message $\mathcal{M}_{i}$ and 
is calculated using the neighborhoods of node $v_j \in \mathcal{N}_i$.
Results will then be accumulated and passed through a non-linear function $\sigma$ such as ReLU.

\subsubsection{Modeling with Relational Data.}
However, the GNN model presented above can only handle homogeneous graph, as it cannot distinguish one relation from others. 
Relational Graph Convolutional Network (RGCN) extents the idea of GNN to relational data by aggregating the weights of different relations in a knowledge graph~\cite{Schlichtkrull-2017}.
The message-passing function of RGCN is defined as:
\begin{equation}
	h_{i}^{(l+1)}=\sigma\left(\sum_{r \in \mathcal{R}} \sum_{j \in \mathcal{N}_{i}^{r}} \frac{1}{c_{i, r}} W_{r}^{(l)} h_{j}^{(l)}+W_{0}^{(l)} h_{i}^{(l)}\right), \label{eq-rgcn}
\end{equation}
where $ \mathcal{N}_{i}^{r} $ is the set of neighbor indices of node $v_i$ under relation $r \in \mathcal{R} $.  $c_{i, r} $ is a normalization constant that can be learned or predefined.

The rapid growth of relations number $\mathcal{R}$ might lead to overfitting on rare relations or to models with enormous amount of parameters.
To prevent that from happening,  model weights $W_r^{(l)}$ regularization is enforced on top of RGCN. 
Authors of RGCN proposed several methods based on the principle of parameters sharing. 
Instead of learning separated parameters for each relations, RGCN learns a group of shared parameters which can then be composited as weights of relations. 
Since the shared parameters is trained by all relations, it would less likely overfit to a specific relation.

For example, one of the methods used for such purpose is basis-decomposition that defines each weight as follows: 

\begin{equation}
	W_{r}^{(l)}=\sum_{b=1}^{B} a_{r b}^{(l)} V_{b}^{(l)}, \label{eq-basis_decomp}
\end{equation}
where each $W_{r}^{(l)}$ is a linear combination of basis transformation $V_{b}^{(l)} \in \mathcal{R} ^{d^{(l+1)} \times d^{(l)}}$ with coefficients $a_{rb}^{(l)}$. In such way, only the coefficients depend on the relation $r$ and therefore prevents model from overfitting and overgrowing. 
This method has been prove to be effective in entity classification tasks~\cite{Schlichtkrull-2017}. 

\section{Federated Relational Graph Modeling}\label{sec-fedrgcn}
As real-world graphs differ significantly, typical FL algorithms, such as FedAVG, cannot be directly used for GNN models without aligning models in each knowledge base.
Existing graph models that adapt to the federated setting is less explored, and most of the existing works can only handle homogeneous graphs \cite{Suzumura-2019,zheng2020asfgnn}.
To facilitate modeling heterogeneous graphs over the federated networks, in this paper, we are the first to propose a Federated version of RGCN, i.e., Fed-RGCN. 

\subsection{Proposed Architecture}
\begin{figure}[tbp]
  \centering
	\includegraphics[width=\linewidth]{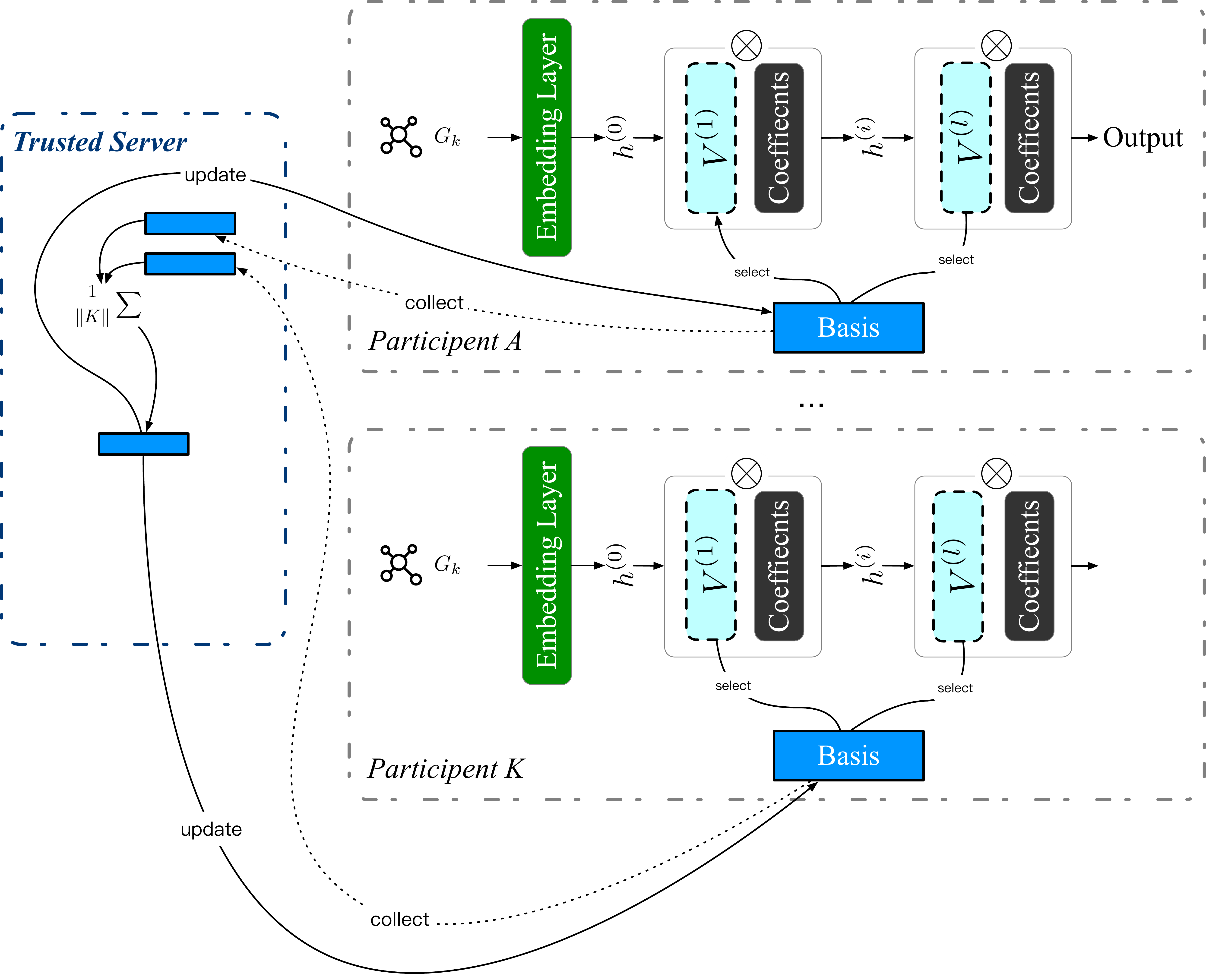}
	\caption{Architecture of Federated RGCN. \label{fig-fedrgcn-arch}}
\end{figure}
As discussed in the last section, to regularize the weights of hidden layers from overfitting and exponentially increasing, RGCN utilizes weight sharing method to map increasing relational weights to the predefined basis.
Such a basis design naturally brings convenience for extending the RGCN to the federated setting. 
As shown in Fig.\ref{fig-fedrgcn-arch}, all the participants build their RGCN models with the basis of the same shape, i.e., with the same dimensions $B$ and number of layers $L$. In each iteration, the global server chose $K$ devices, each updates the model using its own data. Then the server collects their gradients of the basis and aggregating them using a weighted average:
\begin{equation}
  \centering
  \nabla w \leftarrow \frac{1}{\|K\|} \sum_{k=1}^{K} \frac{\nabla w_k}{\|N\|},
  \label{eq-grad-aggregate}
\end{equation}
where $\|N\|$ is the amount of data in each base, which can be calculated in multiple ways. In this paper, we define it as the number of nodes. 
After gradients aggregation, each participant updates the local model with the global one and continues with the local training.

\subsection{Problems of Fed-RGCN}\label{sec-problems}
Although the proposed Fed-RGCN can help tackle the model heterogeneity issue in federated relational data modeling,
using the traditional FL optimization methods such as federated averaging on Fed-RGCN may lead to
slow convergence and degraded performances.
In this section, we analyze two important factors that may affect the federated model convergence.

We first introduce some definitions and assumption for analyzing the convergence of federated learning algorithms~\citet{Li-2020} . Note, $k$ in the following equations denotes the local device and $t$ the current state. 
\begin{definition}{($\gamma^t_k$-inexact solution)}
  Considering a function $h_{k}(w ; w_t)=F_{k}(w)+\frac{\mu}{2}\left\|w-w_{t}\right\|^2$ and $\gamma \in[0,1]$, if there exist a $w^{*}$ such that $\left\|\nabla h_{k}\left(w^{*} ; w_{t}\right)\right\|  \leq  \gamma_{k}^{t} \|\nabla h_{k}(w_{t} ; w_{t})\|$, we call $w^{*}$ is a $\gamma_{k}^{t}-$ inexact solution of $\min_{w} h_{k}\left(w ; w_{t}\right)$.
\label{def-gamma-inexact}
\end{definition}
Since $h_{k}(w_t ; w_t) = F_{k}(w_t)$, $\min_{w} h_{k}(w)$ is actually a subproblem of $\min_w F_k(w)$, whose inexactness of its solutions is bounded by $\gamma_{k}^{t} \in [0, 1]$ if and only if $h_k(w)$ is $1$-Lipschitz continuous.

\begin{definition}{($B$-local dissimilarity)}
  Denote $f(w)$ as the global objective and $F_k(w)$ the local counterpart on $k$ device, the local functions $F_{k}$ are $B$-locally dissimilar at $w$ if $\mathbb{E}_{k}\left[\left\|\nabla F_{k}(w)\right\|^{2}\right] \leq$ $\|\nabla f(w)\|^{2} B^{2} .$ We further define $B(w)=\sqrt{\frac{\mathbb{E}_{k}\left[\left\|\nabla F_{k}(w)\right\|^{2}\right]}{\|\nabla f(w)\|^{2}}}$
for $^{2}\|\nabla f(w)\| \neq 0$.
\label{def-b-local}
\end{definition}

An optimal weight $w$ that minimizes local objective can also minimize the global one if and only if $f(w)$ and $F_k(w)$ is close enough. $B$-local dissimilarity measures such similarity. 
Assume following assumption holds:
\begin{assumption}{(Bounded dissimilarity)}
  For some $\epsilon>0$ and all the points $w \in \mathcal{S}_{\epsilon}:=\left\{w \mid\|\nabla f(w)\|^{2}>\epsilon\right\}$, there exists a $B_{\epsilon}$ such that $B(w) \leq B_{\epsilon}$.
  \label{assumption-b-similarity}
\end{assumption}

With those definitions and assumption,  a federated algorithm is guaranteed to converge in finite iterations. We direct the reader to \citet{Li-2020} for a detailed proof. 
Given a local objective in the form of $h_k(w)$, since $w_t$ will be assign to $w$ in every epoch and $\nabla h_{k}\left(w ; w_{t}\right)=\nabla F_{k}(w)+\mu\left(w-w_{t}\right)$, if Assumption \ref{assumption-b-similarity} holds for $F_{k}(w)$, it holds for $h_k(w)$ as well. 
Hence the dissimilarity between $h_k$ and $f_k$ is bounded, and the corresponding solutions is also bounded by $\gamma_k^t$, indicating there exists a solution $w^*$ of $h_k(w)$ which is close enough to the solution of $f(w)$.

The above-mentioned assumptions and analysis are most likely held for batched samples, e.g., texts or images, 
however, it can hardly be satisfied in the context of graph data modeling, due to the potential divergence and non-smoothness of the objective functions.

\subsubsection{Objective Divergence}\label{sec-objective-div}
Intuitively, the fundamental difference between batched samples and the graph data is their separability. 
While the samples can be easily divided into several mini-batches in any combination, graph can only be separated in strict conditions. 
Considering an ideal graph $G^*$ contains all private graphs $G_k$ in each base.
By aggregating local trained weights $w_k$ trained on each $G_i$, the expectation $\mathbb{E}[w]$ equals the stationary solution $w^{*}$ trained on $G^*$ if and only if each base is separated from the ideal graph by a cut vertex, which means they can form a complete and exclusive set. 
Since the federated network is formed in a stochastic manner, this condition will unlikely be met. 

Therefore, in the same form as batched data modeling, the local objective is defined as $F_k(w) := \mathbb{E}_{G_k \sim \mathcal{M}_k}[f(w;G_k)]$, where $\mathcal{M}_k$ is the set of all possible sub-graphs with $n_k$ nodes or relations. 
Different from the usual setting where $x_k \sim \mathcal{D}_k$, $G_k$ can be sampled only once in each device. 
Such restriction makes the empirical measurement of objective used in practice as just a surrogate of the expectation $\mathbb{E}_{k}\left[\left\|\nabla F_{k}(w)\right\|^{2}\right]$ with significant variance, and thus the $B$-dissimilarity measurement. 
Consequently, the bounded assumption $B(w) \neq B_{\epsilon}$ can hardly be guaranteed, neither can the convergence of federated learning algorithm.

\subsubsection{Objective Unsmoothness}
Another problem that affects the federated algorithm convergence is the smoothness of the objective for graph data modeling.  
Consider the global loss function $f(w; G)$ and $G_x, G_y \in \mathcal{M}_k$, the $L$-Lipschitz smoothness requires $\|f(w;G_x) - f(w;G_y) \| \leq L \| G_x - G_y\|$. 
However, known as the Lipschitz extension problem \cite{Aronsson-1967}, whether a $L$-Lipschitz continuous function $f(\cdot)$ applying on two graphs fulfills the Lipschitz condition depends. It has been proved that Lipschitz extension of higher-dimensional functions on graphs do not always exist \cite{Raskhodnikova-2016}. Therefore, the global objective for federated graph modeling might not $L$-Lipschitz continuous, neither its expectation on the local device $F_k(w)$. Reviewing the definition of $\gamma_k^t$-inexact solution, it is obvious that if $F_k$ and the corresponding $h_k$ are not $L$-Lipschitz continuous, then there 
may not exist a solution $w^*$ that makes federated algorithm to converge.

\section{Proposed Solutions}\label{sec-solutions}
As analyzed in the last section, the challenges of applying federated algorithms on Fed-RGCN rising from the potential divergency  and  non-smoothness of the objective functions.
In this section, we propose a federated learning algorithm, called FedAlign, that utilizes optimal transport to regularize the model divergence and a weight penalty to enforce the objective to be quasi-Lipschitz continuous.

\subsection{Basis alignment\label{sec-penalties}}
As mentioned in section~\ref{sec-objective-div}, divergence between the empirical local objective $F_k(w)$ and the expectation violates the Bounded dissimilarity, which makes the convergence of federated learning algorithm unguaranteed. Using $h_k(w) := F_k(w) + \frac{\mu}{2}\|w - w_t\|^2$ to replace $F_k(w)$ as local objective can alleviated such problem, since the impact of biased $F_k(w)$ can be balanced by penalizing the difference between the local weights and the global ones~\cite{Li-2020}. 
However, such a solution does not work for Fed-RGCN as expected. because we only extract parts of $w$, i.e., the basis for aggregation, which makes $\|w - w_t\|$ not guaranteed to converge towards zero, and thus $h_k$ a biased approximation of $F_k(w)$.



To alleviate this problem, we can view the weights of Fed-RGCN as a sample drawn by a distribution.
Assuming there is a stationary solution of weight $w^*$ who are drawn from certain distribution and the global weight $w$ is an unbiased estimation of $w^*$, we can expect the distance between the distribution of local and global weights converge to $0$. 

\textbf{Optimal Transportation (OT) distance} \cite{Villani-2008} is a widely used measurement for such purpose. Intuitively, OT distance can be viewed as the minimum amount of mass needed to be transferred if we want to turning one pile, which is a distribution defined on a given metric space $M$ into other.
It is also known as earth mover's distance (EMD) \cite{Rubner-1997} in computer vision with the same analogy.
Comparing with other metrics, such as Euclidean distance or Kullback-Leibler divergence, OT distance has some nice properties that make it more suitable for comparing distribution related to graph data. For example, it does not assume compared distributions to be in the same probability space, and unlike KL-divergence, OT distance is symmetric for two distributions.

Different choices of cost function leads to different OT distances. In its simplest form, the cost of a move is the distance between the two points, thus, the OT distance is identical to the definition of the Wasserstein-1 distance or namely the EMD.
Formally, the EMD can be defined as follows.
Given two probability vectors $r$ and $c$, each has a dimension of $n$ and $m$, respectively. Let $U(r,c)$ be the set of positive $n \times m$ matrices, in which the rows sum to $r$ and the columns sum to $c$, we have:
\begin{equation}
 U(r,c) = \{ P \in \mathbb{R}_{+}^{n \times m} | P \mathbf{1}_m=r, P^T \mathbf{1}_n=c \},
\end{equation}
where $\mathbf{1}_m$ is the $m$ dimensional vector of ones.  
 
For  two multinomial random variables $X$ and $Y$ taking values in $\{1, \dots, n\}$ and $\{1, \dots, m\}$, each with distribution $r$ and $c$ respectively, any matrix $P \in U(r, c)$ can then be identified with a joint probability for $(X, Y)$ such that $p(X=i, Y=j)=p_{i j}$. Given a $n \times m$ cost matrix $M$, in which $M_{ij}$ is the cost to move $X=i$ to $Y=j$. The definition of EMD will be:
\begin{equation}
d_{M}(r, c):=\min _{P \in U(r, c)}\langle P, M\rangle,
 \label{eq-ot-defination}
\end{equation}
where $d_{M}(r, c)$ can be solve via linear programing. 

To lower the cost of calculating OT distance, we use Sinkhorn distance \cite{Cuturi-2013} to replace Wasserstein distance. Sinkhorn distance modifies the objective function $d_M(r, c)$ of Wasserstein distance by adding a entropy constraint:
\begin{equation}
d_M^{\lambda}(r, c) = \min _{P \in U(r, c)}\langle P, M\rangle
- \frac{1}{\lambda} h(P),
	\label{eq-sinkhorn}
\end{equation}
where $h(P) = -\sum_{i, j=1}^{d} p_{i j} \log p_{i j}$ and $\lambda \in [0, +\infty)$. The Sinkhorn distance can be calculated via iteratively scaling the rows and columns of $P$. The cost of computing
Sinkhorn distance is $O(d^2)$, while the complexity for calculating Wasserstein distance is at least $O(d^3 \log(d))$. 
We direct reader to \citet{Cuturi-2013} for further reading.

Although the graph models naturally differ due to the inherent heterogeneity of graph data, our proposed Fed-RGCN only needs to aggregate the basis of $V$.
Thus, we only need to calculate the OT distance of basis from different bases. The proximal term that measures the difference between local and global weights 
is then formulated as the average OT distances between the basis in each layer:
\begin{equation}
  \frac{\mu}{\|N\|} \sum_{j \neq k}^{N} \sum_{l=1}^{L} OT(V^{(l)}_k, V^{(l)}_j),
  \label{eq-basis-alignment}
\end{equation}
where $N$ is the number of selected devices and $\mu$ is a hyper-parameter.

\subsection{ Weight Penalty }
To improve the algorithm convergence, we further add a weight penalty to make the objective function quasi-Lipschitz continuous.
Following the previous work~\cite{Gulrajani-2017}, 
we add a weight penalty into loss function:
\begin{equation}
  \lambda (\|\nabla_{G_k}F_k(w)\|_2 - 1)^2,
  \label{eq-weight-penalty}
\end{equation}
where $\lambda$ is a hyper-parameter to be tuned. 

Essentially, this term penalizes the $2$-norm of gradients larger than $1$.
Originated from Wasserstein generative adversarial network (WGAN)~\cite{Arjovsky-2017}, researchers find it is necessary to constrain critic function to $1$-Lipschitz.
Further work by \citet{Gulrajani-2017} shows that applying a soft constraint, i.e., the gradient penalty (GP), is more effective than using hard weight clipping.
Note, the original weight penalty is an expectation calculated using $\hat{x} \sim \mathbb{P}_{\hat{x}}$ for batched samples, the term in \eqref{eq-weight-penalty} can only perform in the whole graph since we can not split the graph in the current federated setting. This may cause the penalty biased to local data, a further improvement introducing $G_k \sim \mathcal{M}_k$ into it will be favored. 


\subsection{Algorithm}
Combining the basis alignment and weight penalty, the local loss function of federated RGCN is defined as:  
\begin{equation}
\begin{aligned}
	\hat{F}_k(w)&:=F_k(w) \\
	& + \frac{\mu}{\|N\|} \sum_{j \neq k}^{N} \sum_{l=1}^{L} OT(V^{(l)}_k, V^{(l)}_j) \\
	& + \lambda (\|\nabla_{G_k}F_k(w)\|_2 - 1)^2.
\end{aligned}
\label{eq-fedrgcn-loss}
\end{equation}

In Algorithm \ref{algo-fedalign}, we present the optimization process for the federated relational data modeling. The resulting algorithm is referred to as FedAlign.
Here, $\mu$ and $\lambda$ are the hyper-parameters that control the basis alignment and weight penalty.
$k=1, \dots, N$ denotes $N$ devices that participate in the federated training.
$E_{local}$ denotes the number of epochs trained for each local device before it sends its gradients to serve and $E_{global}$ denotes the number of epochs
for the whole training process.
Note, a stochastic gradient descent (SGD) optimizer with fixed learning rate $\alpha$ is used in our implementation, however, other optimizers such as Adam \cite{Kingma-2017} can also be used.
\begin{figure}[tbp]
 \removelatexerror
  \begin{algorithm}[H]
    \SetKwInOut{Input}{Input}
    \SetKwInOut{Output}{Output}
    \SetKwInOut{Server}{Server}
    \SetKwInOut{Client}{Client}
    \SetAlgoLined
    \caption{FedAlign (Proposed Algorithm)}
   \Input{$\mu, \lambda, \nabla w^{0}, N, E_{global}, E_{local}, k=1, \cdots, N, SGD(\alpha)$}
\Server{}
Set current global epoch $e=1$\;
Send initial gradient of basis weights $\nabla w^{0}$ to clients\;
 \While{$e \leq E_{global} $}{
  Collecting gradients $\nabla w_k$ sent by client $k$\;
  \If{$\|\nabla w_k\| = N$}{
   Aggregating $\nabla w \leftarrow \frac{1}{\|N\|}\sum_{k=1}^{N}{\nabla w_k}$ \;
   Send $\nabla w_k$ to each clients\;
   }
   Set $e = e + 1$\;
 }

 \Client{Client $k \in N$ side with graph $G_k$}
 Initialize basis weigh by $w$ received from Server\;
   \For( \emph{local update}){$e := 1$ to $E_{local}$}
   {
      Calculating loss $\hat{F}_k(w)$ according to \eqref{eq-fedrgcn-loss} \;
      Collected gradients $\frac{\nabla \hat{F}_k}{\nabla w_k}$ of SGD optimizer\;
      Update local weight $w_k \leftarrow w_{k}^{t-1} + \alpha$ $\frac{\nabla \hat{F}_k}{\nabla w_k}$\;
   }
   Send gradients collected to Server\;
   \KwResult{ $w \leftarrow \frac{1}{\|N\|}\sum_{k=1}^{N} w_k$ }
     \label{algo-fedalign}
  \end{algorithm}
\end{figure}


\section{Empirical evaluation}\label{sec-empirical-evaluation}
We evaluate the proposed algorithm with Fed-RGCN on entity classification task to verify its performance. Six settings are studied in the experiments with three federated algorithms, i.e., FedAVG, FedProx, FedAlign, and their variants with weight penalty (denoted by -L).

\subsection{Synthetic Datasets}

We use three commonly used public datasets in the Resource Description Framework (RDF) format: AIFB, MUTAG, and BGS \cite{Ristoski-2016} to test the performance of the proposed algorithm. 
The dataset contains different types of entities and relations, as shown in Table \ref{tab-dataset}.
\begin{table}[htbp]
  \centering
  \begin{tabular}{ccccc}
  \hline
  {}        & \textbf{Types} & \textbf{Entities} & \textbf{Relations} & \textbf{Edges}  \\
  \hline
  \textbf{AIFB}  & 7     & 8285     & 104       & 29043 \\
  \textbf{MUTAG} & 5     & 23644    & 50        & 74227 \\
  \textbf{BGS}   & 27    & 333845   & 122       & 916199 \\
  \hline
  \end{tabular}
  \caption{The details of each dataset.\label{tab-dataset}}
\end{table}

A specific type of entity has been labeled to be used as the classification target. 
The dataset provider has split them into two sets for training and testing. 
The number of classes and size of both sets can be seen in Table \ref{tab-datasetsplit}.
\begin{table}[]
  \centering
  \begin{tabular}{cccc}
      \hline
  {}        &  \textbf{Classes} & \textbf{Train Set} & \textbf{Test Set} \\ 
  \hline
  \textbf{AIFB} & 4       & 140      & 36      \\
  \textbf{MUTAG} & 2       & 272      & 68      \\
  \textbf{BGS} & 2       & 117      & 29    \\  \hline  
  \end{tabular}
  \caption{The detail of labeled entities in each dataset.\label{tab-datasetsplit}}
  \end{table}

To mimic the setting of federated knowledge bases, 
we split each dataset into $N=10$ parts in the following way.
First, for nodes that are not labeled, we randomly select $6$ types (excepts MUTAG for $5$) and sample $\|N^{(i)}_{ntype}\| \sim U(0, \|N_{ntype}\|)$ nodes from the complete dataset for each base. We then shuffle the labeled nodes in the training set and split it into $N$ parts, each for a client. The labeled nodes in the test set will be duplicated and stored in each client, but keep unused during the training process.
Finally, we add an edge from the complete dataset into a base if it contains its source and destination nodes. 
The number of nodes and edges in each base are listed in \ref{tab-fed-dataset}.
\begin{table}[htbp]
\centering
\begin{tabular}{ccc}
      \hline
{}        & \textbf{Entities(Each)}  & \textbf{Edges (Each)}\\
\hline
\textbf{AIFB}  & 2993.00 $\pm$ 1737.77     & 7923.60 $\pm$ 7092.13\\
\textbf{MUTAG} & 6537.10 $\pm$ 2634.75     & 4578.20 $\pm$ 1899.34\\
\textbf{BGS}   & 6123.40 $\pm$ 5667.58     & 4671.00 $\pm$ 5930.07 \\      \hline
\end{tabular}
\caption{The mean number and variance of entities and edges in synthetic federated dataset.\label{tab-fed-dataset}}
\end{table}

As we can see from Table \ref{tab-fed-dataset}, although we choose the same numbers of types and relations for each base, the size of entities and edges can still differ tremendously. Such a phenomenon is caused by the unbalanced distribution of entities in different types, and also the vanish of edges if its source and destination are in different bases. 
It indicates that, in the federated setting of relational data modeling, even with a balanced setup, the statistical heterogeneity of dataset can still be significant.

\subsection{Implementation}
We implemented the FedAlign, FedAVG and FedProx on RGCN models to compare the algorithm performances.
The RGCN model is constructed following the previous work \citet{Schlichtkrull-2017} with $2$ hidden layers and a constant number of basis $\|V\|$. 
Both three federated algorithms are optimized via a SGD optimizer \cite{Bottou-1991} whose learning rate is $\alpha$. Note, the $\|V\|$ and $\alpha$ are the  hyper-parameters needs to be tuned. 

Algorithms are mostly implemented using Pytorch \cite{Paszke-2017} and DGL \cite{Wang-2019c} library.
Sinkhorn algorithm is implemented with Geomloss \cite{Feydy-2019}. 
We also use Tune \cite{Liaw-2018} to grid search the hyper-parameters.

\subsection{Hyper-parameters}\label{sec-finetune}

Hyper-parameters settings have significant impacts on the performance of RGCN as well as the federated algorithm.
We focus on tuning four parameters: the number of basis $\|V\| \in [1, 50, 100]$, the learning rate $\alpha \in [0.01, 0.05, 0.1]$,  the factor of basis alignment term $\mu \in [0.1, 1, 10]$, and the factor of weight penalty $\lambda \in [0.1, 1, 10]$.
RGCN and optimizer related parameters, i.e., $\|V\|$, $\alpha$ and $\lambda$,  will affect all three algorithms, while $\mu$ only affects FedProx and FedAlign that constrains the divergence between global and local weights. 
Surprisingly, the optimal hyper-parameters for all six settings is the same, in which $\|V\|=100$, $\alpha=0.1$, $\mu=10$ and $\lambda=10$. While the value of $\alpha$ and $\lambda$ is widely used in practice\citet{Gulrajani-2017}, the value of $\|V\|$ and $\mu$ is very different from existing literature
(in which is $30 \sim 50$ and $0.01$)\cite{Schlichtkrull-2017,Li-2020}. Such difference might be caused by the difference between batched samples and relational data.

In addition, two parameters control the amount of computation, i.e., $E_{local}$ is the number of training makes over the local dataset of each client on each round and  $E_{global}$ denotes the global number of epochs that aggregating all devices. Due to the limited computation resources, we 
set the $E_{local}=5$ and $E_{global}=20$ for all datasets.

\subsection{Results}
\begin{figure*}[tbp!]
  \centering
  \begin{subfigure}{.3\linewidth}
    \includegraphics[width=\linewidth]{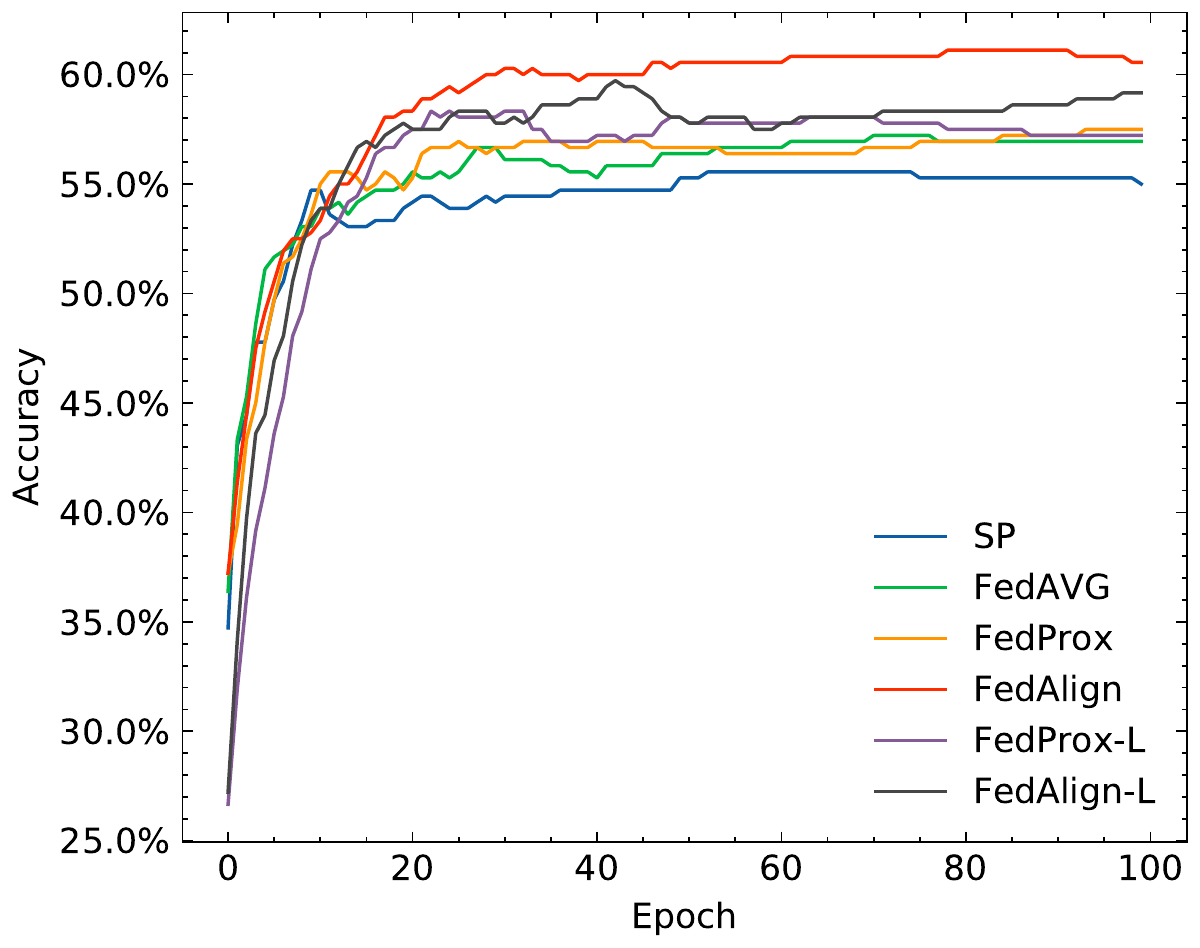}
    \caption{AIFB\label{subfig-aifb}}
  \end{subfigure}
  \begin{subfigure}{.3\linewidth}
    \includegraphics[width=\linewidth]{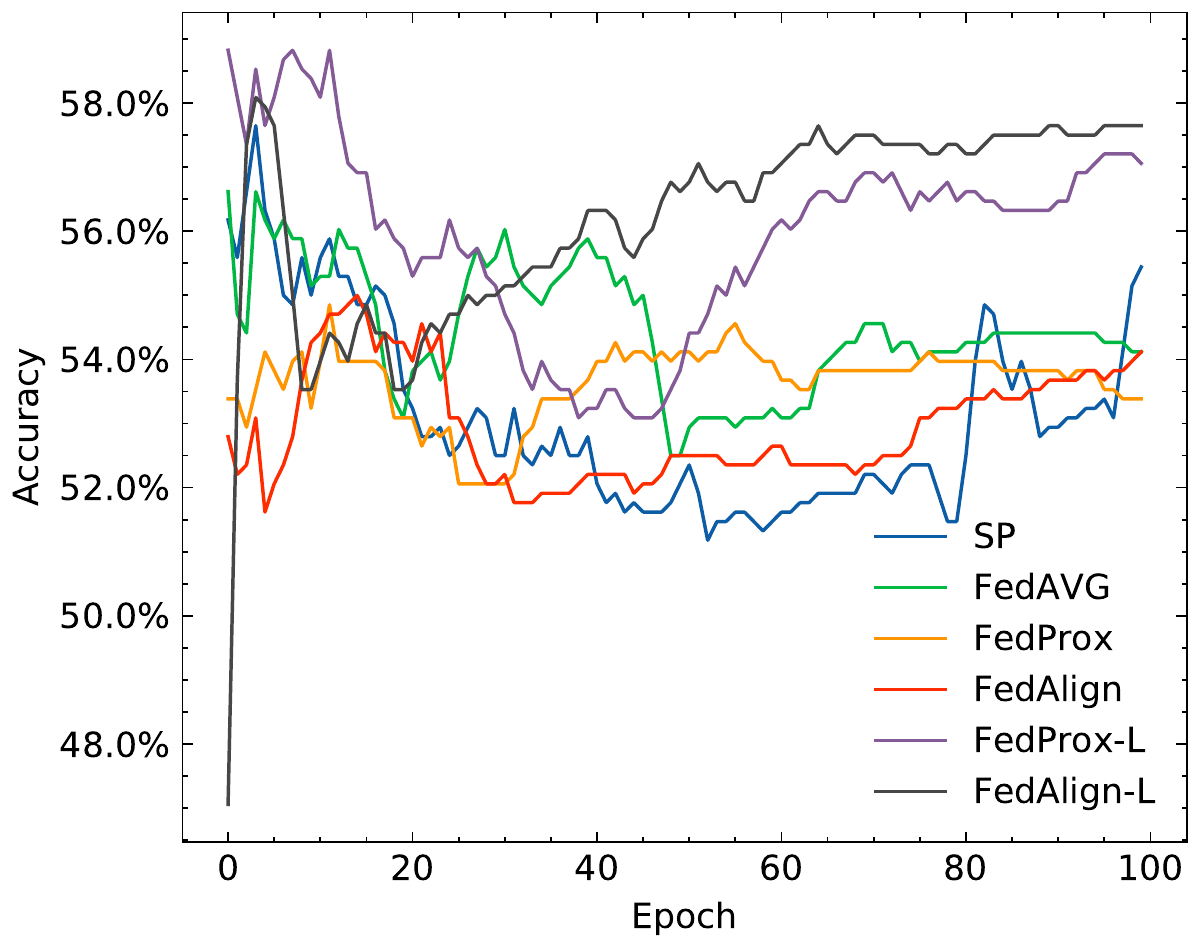}
    \caption{MUTAG\label{subfig-mutag}}
  \end{subfigure}
  \begin{subfigure}{.3\linewidth}
    \includegraphics[width=\linewidth]{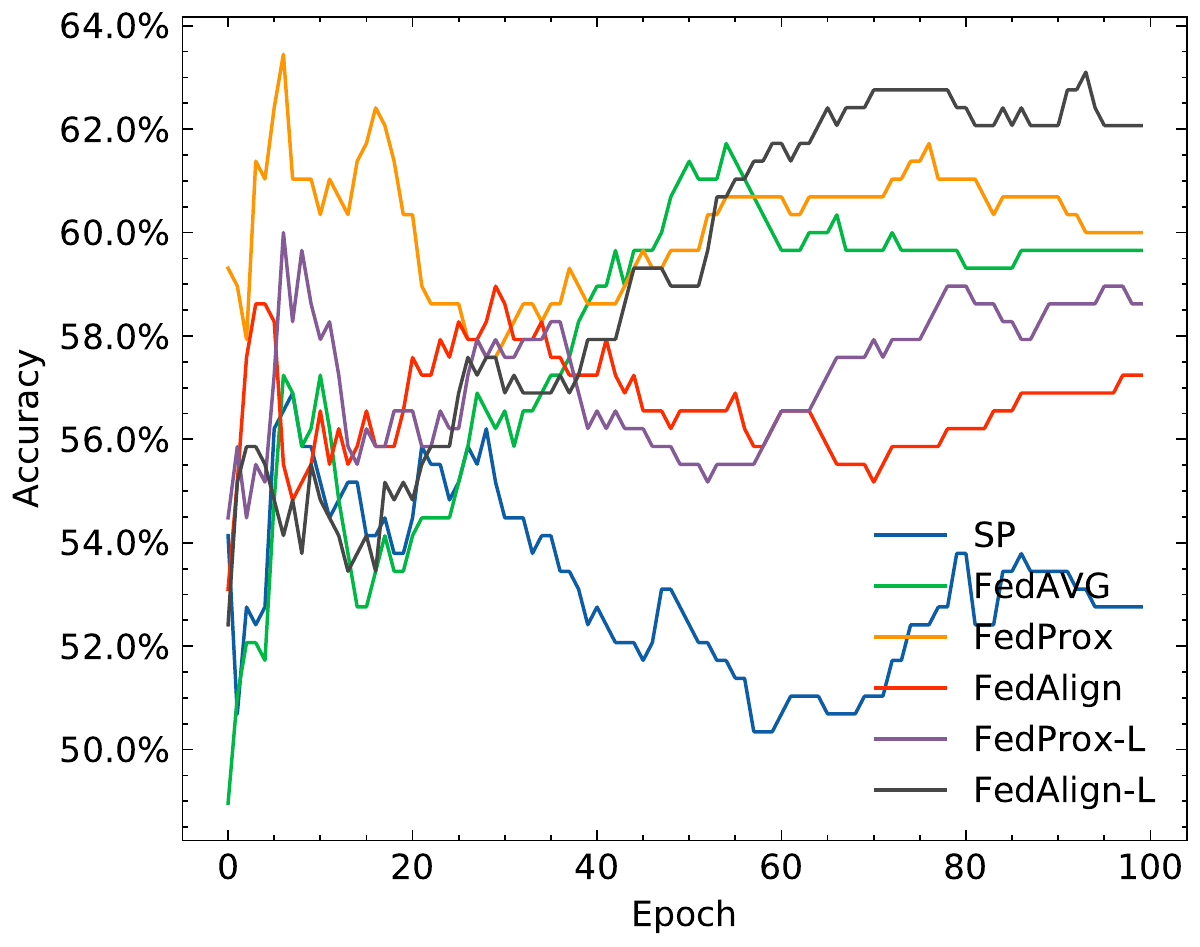}
    \caption{BGS\label{subfig-bgs}}
  \end{subfigure}
  \caption{Accuracy on validation set of different models on each dataset.\label{fig-performance}}
\end{figure*}

For separated learning (SP, i.e., to train only on its own device for each participant), three FL algorithms (i.e., FedAVG, FedProx, FedAlign) and their $1$-Lipschitz regularized variants (i.e., FedAVG-L, FedProx-L and FedAlign-L), we run 10 federated training on the separated datasets, then aggregates weights of each base into a global basis, which will then be synchronized to the local model before evaluation.
Performance results are measured by the classification accuracy and shown in Table \ref{tab-performance}. 

\begin{table}[htbp]
\centering
\begin{tabular}{cccc}
      \hline
 {}       & \textbf{AIFB}                & \textbf{MUTAG}               & \textbf{BGS}                    \\
\hline
\textbf{SP}         & 55.00\%        & 55.44\%        & 52.76\%   \\
\hline
\textbf{FedAVG}     & 56.94\%          & 54.12\%          & 54.86\% \\
\textbf{FedProx}    & 57.50\%          & 53.38\%          & 55.24\% \\
\textbf{FedAlign}   & \textbf{60.56\%} & 55.17\%          & 57.24\%  \\
\hline
\textbf{FedAVG-L}     & 57.94\%          & 54.71\%          & 55.48\% \\
\textbf{FedProx-L}  & 57.22\%         & 57.06\%          & 58.62\%   \\
\textbf{FedAlign-L} & 59.17\%          & \textbf{57.65\%} & \textbf{60.07\%}    \\      \hline      
\end{tabular}
\caption{Performance of different algorithms on four dataset.\label{tab-performance}}
\end{table}

\subsubsection{Basis Alignment}
As we can see from Table~\ref{tab-performance}, FedAlign outperforms other federated algorithms on all three datasets. Comparing with FedAVG and FedProx, FedAlign improves the classification accuracy by $1.05\% \sim 3.62\%$ on average. 

We notice that separated training outperforms most of the federated algorithms on MUTAG dataset. 
Interestingly, the original RGCN performs worse than the  traditional methods on MUTAG and BGS datasets as well.
\citet{Schlichtkrull-2017} attributing the problem to the nature of datasets. Since MUTAG is a dataset of molecular graphs and BGS of rock types with hierarchical feature, their relations can either indicate atomic bonds or merely the presence of a certain feature.
Therefore, the labeled entities in them can only be connected via high-degree hub nodes, such as the name of molecular or rock that encodes a certain feature. In other words, the graph structure will most likely be star-shape, and its information are stored in attributes instead of structures. 
Modeling these kind of relations needs understanding of the contents in node attributes or the structure of complete graph. Comparing with methods such as RDF2Vec embeddings \cite{Ristoski-2016a} and Weisfeiler-Lehman kernels (WL) \cite{Shervashidze-2011, deVries-2015}, which captures such information, RGCN uses only randomized embedding and messages from neighborhoods, thus limits the performance of the model.

Such problem could be even worse for federated learning scheme. Comparing with graph connected via a more centralized way, the structure of star-shape network will more likely to be break by the distributed setting. Such situation will cause tremendous information loss. As shown in Table~\ref{tab-fed-dataset}, each base in federated MUTAG contains only $6.1\%$ edges of the complete dataset, and federated BGS only $0.5\%$.
Since the size of dataset could be too small, overfitting to local structure could possibly happened. 

We randomly select one training log that shown in Fig.~\ref{fig-performance}. The performance is evaluate in each global epoch using aggregated global model on test set. It can be seen that, comparing with models trained on AIFB, models trained on MUTAG and BGS suffering overfitting more significantly. Since federated algorithms aggregating parameters collected from each participants, models that overfitting to local dataset will probably undermining the performance of global model.

\subsubsection{Weight Penalty}
$1$-Lipschitz weight penalty can be viewed as an regularization upon model that prevents it from overfitting to local data as analysis in WGAN-GP \cite{Gulrajani-2017}. We observed similar results in our experiments.
As shown in Fig.~\ref{subfig-mutag} and \ref{subfig-bgs}, comparing with original algorithms, those with $1$-Lipschitz penalty, i.e. FedAVG-L, FedProx-L and FedAlign-L have better performances in general. Moreover, for the MUTAG and BGS datasets, FedProx-L and FedAlign-L continuously improve after performance declines in the early stage, while FedProx and FedAlign stay stationary in most of the training stage. The performance of $L$-Lipschitz constrained algorithm 
improved $1\%\sim 5\%$. Such phenomenon indicates that the models have been stuck in local optimal points.

\subsection{Discussion}

Though the proposed algorithm with basis alignment and weight penalty outperforms FedAVG and FedProx on relational data modeling, it should notice that, all the models trained on federated bases are still underperformed by the model training on complete graph 
as reported by \citet{Schlichtkrull-2017}). 
As we analyzed in Section \ref{sec-problems}, the problems underlying in federated data modeling is the non separability of graph data which leads to a divergence of local loss function and global counterpart, and the incomparability leads to the non-Lipschitz condition. The proposed workarounds can alleviate but hardly eliminate them.
Moreover, the information loss, such as edges connected entities in separated bases can not be restored in federated setting. Both problems implies the future work of federated relational data modeling might focus on changing the non-separability and incomparability of graph data. 



\section{Conclusion \label{sec-conclusion}}

We analyzed the problems of existing federated modeling on relational data, and proposed FedAlign algorithm to handle them. By using OT distance to measure the divergences of basis in different models and adding $L$-Lipschitz weight penalty to training process, the accuracy of Fed-RGCN could improve with acceptable extra computational cost. 
Our empirical evaluation has shown the proposed algorithm outperforms state of art methods, such as FedAVG and FedProx on SRL task. 
As far as we are acknowledged, this is one of the earliest attempts to handle knowledge-graph related missions via federated learning. The study of applying privacy-preserving techniques on graph data remains largely untouched. 
There is no widely applied methods for some important problems, such as entities alignments, link prediction, that can be performed without leaking the private information. Such situation limits the usage of relational data and requires a change.
We hope our work could provide useful insight for the community and push the research forward. 

\bibliography{references.bib}

\end{document}